\documentclass[twoside]{article}
\usepackage[accepted]{aistats2015arxiv}

\usepackage{graphicx} 
\usepackage{subfigure}
\usepackage{graphicx} \usepackage{amsmath} \usepackage{amssymb}
\newcommand{\BEQ}{\begin{equation}}
\newcommand{\EEQ}{\end{equation}}
\newcommand{\BEA}{\begin{eqnarray}}
\newcommand{\EEA}{\end{eqnarray}}
\newcommand{\BGA}{\begin{gather}}
\newcommand{\EGA}{\end{gather}}

\newcommand{\e}{\epsilon}

\newcommand{\ua}{\uparrow}
\newcommand{\da}{\downarrow}
\newcommand{\UA}{\Uparrow}
\newcommand{\DA}{\Downarrow}

\newcommand{\CN}{{\cal N}}
\newcommand{\bx}{{\bf x}}
\newcommand{\bhx}{{\bf \hat{x}}}
\newcommand{\bhz}{{\bf \hat{z}}}
\newcommand{\btz}{{\bf \tilde{z}}}
\newcommand{\by}{{\bf y}}
\newcommand{\bz}{{\bf z}}

\newcommand{\btau}{ \boldsymbol\tau  }
\newcommand{\comment}[1]{}

\DeclareMathOperator*{\argmax}{arg\,max}
\begin{document}

%
\runningtitle{Active Inference for Binary Symmetric HMMs}

%

\twocolumn[

\aistatstitle{Active Inference for Binary Symmetric HMMs}

\aistatsauthor{ Armen Allahverdyan  \And Aram Galstyan }
\aistatsaddress{ Yerevan Physics Institute\\Yerevan, Armenia  \And USC Information Sciences Institute \\ Marina del Rey, CA, USA  } 
]



\begin{abstract}
   We consider {\em active}  maximum a posteriori (MAP) inference problem for 
 Hidden Markov Models (HMM), where,  given an initial MAP estimate of the hidden sequence, we select to label certain states in the sequence to improve the estimation accuracy of the remaining states. We develop an analytical approach to this problem for the case of binary symmetric HMMs, and obtain a closed form solution that relates  the expected error reduction to model parameters under the specified active inference scheme. We then use this solution to determine most optimal active inference scheme in terms of error reduction, and examine the relation of those schemes to heuristic principles of  uncertainty reduction and solution unicity. 

\end{abstract}

\section{Introduction}
\label{sec:intro}


In a typical statistical inference problem, we want to infer the true
state of some hidden variables based on the observations of some other
related variables.  The quality of statistical inference is generally
improved with adding more data or prior information about the hidden
variables. When obtaining such information is costly, it is desirable
to select this information optimally so that it is most beneficial to
the inference task.
This notion of active inference, or active
learning~\cite{review_active}, where one optimizes not only over the
models, but also over the data, has been used in various inference
problems including Hidden Markov Models
(HMMs)~\cite{Anderson2005,Hoffman2001}, network
inference~\cite{Zhu2003,Bilgic2009,Bilgic2010,Moore2011}, etc.




The efficiency of active inference is naturally determined from the
error reduction of the inference method \cite{scott}. Since this
quantity is not readily available in practice, several heuristic
methods were developed \cite{review_active}. For instance, a class of
methods seeks to implement active inference in a way of obtaining
possibly unique solution (or reducing the set of available solutions)
\cite{review_active,opper}. A related familiy of methods tries to
learn those variables whose uncertainty is large
\cite{review_active,lewis}. Both heuristics have intuitive rationale
and can do well in practice, but theoretical understanding of how
exactly those heuristics relate to error reduction is largely
lacking. In particular, most existing theoretical results are
concerned with {\em sub-modular} cost functions that allow to
establish certain optimality
guarantees~\cite{Krause2005_UAI,Krause2009_JAIR} on active selection
strategies. However, these cost functions  do not usually refer
  directly to the error reduction of some inference method.

Here we consider the active maximum a posteriori (MAP) inference
problem for the binary symmetric HMMs, or BS-HMM (see
\cite{Anderson2005} for a brief review of active estimation methods in
HMM). BS-HMM is sufficiently simple so it allows to study active MAP
inference analytically \cite{armen_aram,hadler}, yet it still contains
all the essential features of HMMs that make them a flexible and
widely used tool for probabilistic modeling in various fields
\cite{rabiner_review,ephraim_review}. We emphasize  that even for the
  simple BS-HMM model considered here, obtaining analytical insights about active inference strategies is a non-trivial problem, as we demonstrate below.


We develop an analytical approach for studying the active MAP
inference problem for BS-HMM, and use it to determine the most
efficient schemes of uncertainty reduction, where labeling a small
number of states results in the largest expected error reduction in
the estimation of the remaining (unlabeled) states. Our analysis
allows us examine how the active estimation in BS-HMM relates to the
heuristic principles of uncertainty reduction and solution
unicity. Specifically, we obtain a closed form expression for the
expected error reduction as a function of model parameters, and use it
to assess the optimality of active inference schemes.  Finally, we
compare our analytical predictions with numerical simulations and
obtain excellent agreement within the validity range of the proposed analysis. 
 
The rest of the paper is organized as follows: Section~\ref{sec:map} defines
  generalities of active MAP-inference. Sections~\ref{sec:ising} and~\ref{sec:domain} introduce
  the BS-HMM and its MAP analysis, respectively. Section~\ref{sec:active} presents
  our the main analytical  findings on active inference, and is followed by experimental validation of those findings in Section~\ref{sec:numerics}. We conclude by 
  discussing our main findings, their relationships to the existing heuristics for active inference, and  identifying some future directions of research in Section~\ref{sec:conclusion}.

\section{Active inference: generalities}
\label{sec:map}

\subsection{MAP inference}

Let $\bx=(x_1,\ldots,x_N)$ and $\by=(y_1,\ldots,y_N)$ be realizations
of discrete-time random processes ${\cal X}$ and ${\cal Y}$,
respectively. ${\cal Y}$ is the noisy observation of ${\cal X}$.  We
assume a binary alphabet $x_i=\pm 1$, $y_i=\pm 1$ and write the
probabilities as $p (\bx)$ and $p (\by)$.  The influence of noise is
described by the conditional probability $p(\by |\bx )$. Given an
observation sequence $\by$, the MAP estimate
$\hat{\bx}(\by)=(\hat{x}_1(\by),..., \hat{x}_N(\by))$ of $\bx$ is
found from maximizing over $\bx$ the posterior probability
$p(\bx|\by)={p(\by|\bx)p(\bx)}/ {p(\by)}$:
\BEQ 
\hat{\bx}(\by) = \argmax_{\bx}p(\bx|\by).
\EEQ
In general, the minimization can produce $\CN(\by)$ outcomes. Since
they are equivalent, each of them is chosen (for a given $\by$) with
probability $1/\CN(\by)$, which defines a realization $\bhx$ of a
random variable ${\cal \hat{X}}$.


The estimation accuracy is measured using  the 
overlap (larger $O$ means better estimate)
\begin{eqnarray}
  \label{eq:1}
\bar{O}=\sum_{\by}p(\by)\frac{1}{\CN(\by)}\sum_{\bhx (\by)}
O [\bhx(\by),\by], \\
O [\bhx(\by),\by]=\sum_{\bx}\sum_{i=1}^N \hat{x}_i(\by)x_i\, p(\bx|\by),
  \label{eq:1.1}
\end{eqnarray}
where three averages (over ${\cal X}$, ${\cal Y}$ and ${\cal
  \hat{X}}$) are involved in $\bar{O}$, and where the error
probability reads $P_{\rm e}=\frac{1}{2}[1-\frac{\bar{O}}{N}]$. Below
we are interested by the conditional|over a fixed $\by$ and
$\bhx$|overlap $O [\bhx(\by),\by]$. 

\subsection{Active MAP inference}

Given $\by$ and $\hat \bx (\by)$ we request the true value of one of
the variables (say $x_1$, generalization to many
variables is straightforward). Below we refer to this procedure as {\em supervising}. After supervising $x_1$, we
re-estimate other variables:
\begin{eqnarray}
  \label{eq:2}
(\hat{x}_2(\by),..., \hat{x}_N(\by)) \to
(\tilde{x}_2(x_1,\by),..., \tilde{x}_N(x_1,\by)),
\end{eqnarray}
where $\tilde{x}_i(x_1,\by)$ is obtained under the MAP estimation with
the fixed value of $x_1$. Let us define the overlap gain
\begin{eqnarray}
     \Delta  O[\bhx(\by),\by;1]=\sum_{\bx}\sum_{i=2}^N
[\tilde{x}_i(x_1,\by)-\hat{x}_i(\by)]
x_i\,p(\bx|\by),
\label{eq:3}
\end{eqnarray}
Then supervising $x_1$ is meaningful if this gain is positive, $\Delta O >0$. Note that the sum $\sum_{i=2}^N$ is taken over the non-supervised
variables [cf. (\ref{eq:1})]. In $\Delta O [\bhx(\by),\by]$ the
average is taken also over all possible values $\bx=(x_1,...,x_N)$.
The fully averaged overlap gain $\Delta\bar{O}$ is determined from 
$\Delta O [\bhx(\by),\by;1]$ as in (\ref{eq:1}).

If $x_1$ was already correctly recovered using the initial MAP estimation, then supervising $x_1$ does not change anything:
$ \{\tilde{x}_i(\hat{x}_1(\by),\by)\}_{i=2}^N
=\{\hat{x}_i(\by)\}_{i=2}^N$.
Hence,  we get from (\ref{eq:3}):
\begin{eqnarray}
  \label{eq:33}
  \Delta  O[\bhx(\by),\by;1]= \sum_{x_2,...,x_N}\sum_{i=2}^N
[\tilde{x}_i(-\hat x_1(\by),\by)-\hat{x}_i(\by)] \nonumber \\
\times  x_i\, p(-\hat x_1(\by),x_2,...,x_N|\by).
  \label{eq:333}
\end{eqnarray}

\section{Reducing HMMs to the Ising model }
\label{sec:ising}


We now consider a binary symmetric HMM.
${\cal X}$ is a Markov process given by the following state-transition probabilities
\BEA
\label{orleans}
p(\bx)={\prod}_{k=2}^N\, p(x_{k}|x_{k-1})p(x_1),
\qquad x_k=\pm 1,
\EEA
where $p(x_{x}|x_{k-1})$ is the transition probability 
parameterized by a single number $0<q<1$:
\BEA
p(+1|+1)&=&p(-1|-1)=1-q, \nonumber \\
p(+1|-1)&=&p(-1|+1)=q. \nonumber
\EEA


The noise is memory-less and unbiased:
\begin{gather}
\label{bejrut}
p(\by|\bx) = {\prod}_{k=1}^N \pi (y_k|x_k), \qquad y_k=\pm 1
\end{gather}
where
\BEA
\pi(-1|+1)&=&\pi(+1|-1)=\epsilon, \nonumber \\
\pi(+1|+1)&=&\pi(-1|-1)=1-\epsilon,  \nonumber
\EEA
and $\epsilon$ is the probability of
error. Here memory-less refers to the factorization in (\ref{bejrut}),
while unbiased means that the noise acts symmetrically on both
realizations of the Markov process: $\pi(1|-1)=\pi(-1|1)$.  

Recall that the composite process ${\cal X}{\cal Y}$ is Markovina, but ${\cal Y}$ is
generally not a Markovian process~\cite{ephraim_review,rabiner_review}. 

To proceed further, we use the following parametrization of the HMM~\cite{zuk}:
\begin{gather}
\label{om}
p(x_{k}|x_{k-1})=\frac{e^{J x_{k} x_{k-1} }}{2\cosh J} , ~~~~
J=\frac{1}{2}\ln\left[\frac{1-q}{q } \right], \\
\pi (y_{i}|x_{i})=\frac{e^{h y_i x_i}}{2\cosh h}, ~~~~
h=\frac{1}{2}\ln \left[
\frac{1-\epsilon}{\epsilon}
\right],
\label{kaskad}
\end{gather}
MAP estimation is then equivalent to minimizing the  energy function (Hamiltonian) defined as $H(\by,\bx)\equiv -\ln [\, p(\by|\bx)p(\bx)\,] $ \cite{zuk}:
\BEA
\label{jershalaim}
H(\by,\bx)=-J{\sum}_{k=1}^N x_{k} x_{k+1}  - h{\sum}_{k=1}^Ny_kx_k, \EEA
 where we have omitted irrelevant factors that are either additive or
vanish for $N\gg 1$.

We find it convenient to introduce the following {\em gauge} transformation: 
\begin{eqnarray}
  \label{eq:5}
z_i=x_iy_i, \qquad \tau_i=y_iy_{i+1}.
\end{eqnarray}
For a given $\by$, the role of observations will be played
by ${\btau}=\{ \tau_i \} $, while the original hidden variables now correspond to 
$\bz=\{ z_i \}$. Hence, (\ref{jershalaim}) reads
\begin{eqnarray}
  \label{judea}
  H(\btau,\bz)&=&-J{\sum}_{k=1}^N\tau_k z_{k} z_{k+1}
  -h{\sum}_{k=1}^Nz_k,\\
  p(\btau,\bz)&\propto& \exp[-  H(\btau,\bz)].
\label{samarea}
\end{eqnarray}
We shall refer to $\tau_k=\pm 1$ as bond (plus or minus), and to
$z_k=\pm$ as spin (up and down). Eq.~(\ref{judea}) describes the Hamiltonian of a  
one-dimensional Ising model with random bonds $\tau_k$ in an external
field $h$ \cite{zuk}. The MAP inference reduces to minimizing
  this Hamiltonian for a given $\btau$, i.e. to taking the
  zero-temperature limit $T\to 0$ \cite{landau}. Note also that Eq.~(\ref{samarea})
is the Gibbs distribution at temperature $T=1$~\cite{landau}. 

In the new variables, the overlap gain given by Eq.~(\ref{eq:333}) reads:
\begin{eqnarray}
      \Delta  O[\bhz,\btau;1]=\sum_{z_2,...,z_N}\sum_{i=2}^N
[\tilde{z}_i(-\hat z_1(\btau),\btau)-\hat{z}_i(\btau)] \nonumber \\  z_i\, 
p(-\hat z_1(\btau),z_2,...,z_N|\btau).
  \label{eq:7}
\end{eqnarray}
Below we suppress the arguments of $\Delta O$, if they are clear from
the context.

\section{Domain structure of MAP}
\label{sec:domain}

  The MAP estimation for BS-HMMs is characterized by (countably
  infinite) numbers of operational regimes that are related to each
  other via first-order phase transitions~\cite{armen_aram}. Those
  transitions occur at the parameter values 
  \begin{eqnarray}
    \label{eq:6}
    h_c^{m}=2J/m, 
  \end{eqnarray}
  or alternatively, at the critical noise
  intensities~\cite{armen_aram} 
\BEQ \epsilon_c^{m}=1/[1+\e^{4J/m}] \
  , \ m=1,2,..
  \label{eq:6.b}
  \EEQ 

Furthermore, for any noise intensity above the critical value
  $\epsilon >\epsilon_c^1$, the MAP solution is macroscopically
  degenerate: For any observation sequence of length $N$, there are
  {\em exponentially many} (in $N$) sequences that simultaneously
  maximize the MAP objective~\cite{armen_aram}.

  Let us focus on the structure of the MAP solution. A straightforward
  analysis (first carried out in \cite{williams} for the
  one-dimensional Ising model) shows that the dependence of $\hat\bz
  (\btau)$ on $\btau$ is local: $\btau$ is factorized into
  non-overlapping domains $\btau=(\btau_1,{\bf w}_1,\btau_2,{\bf
    w}_2,...)$, where ${\bf w}_k$ are the walls separating
  domains. Walls are sequences of two or more up spins joined by
  positive bonds; see below. The estimate sequence $\hat\bz$ admits
  similar factorization $\hat\bz=(\hat\bz_1,{\bf w}_1,\hat\bz_2,{\bf
    w}_2,...)$, such that $\hat\bz_k$ has the same length as
  $\btau_k$, and is determined solely from $\btau_k$ (non-uniquely in
  general).

  We now proceed to describe the domain structure for those different
  regimes as characterized by $h/J$. Without loss of generality we
  assume $J>0,\, h>0$.

\subsection{The maximum likelihood regime}

For sufficiently weak noise [cf. (\ref{kaskad})]
\begin{eqnarray}
  \label{eq:9}
  h>2J,
\end{eqnarray}
the MAP estimate copies observations $\hat\bx=\by$. Hence, $\hat z_i=1$
for all $i$; see (\ref{judea}). The prior probability $p(\bx)$ is
irrelevant in this regime: on can take $J=0$, i.e. the MAP estimation
reduces to the maximum-likelihood estimation, where the noise
probability $p(\by|\bx)$ is maximized for a fixed $\by$; see
(\ref{bejrut}).

The overlap gain (\ref{eq:7}) nullifies for (\ref{eq:9}), because
supervising a spin does not change other spins: $\hat z_i=\tilde z_i$.

\subsection{First non-trivial regime}

We now focus on the following regime:
\begin{eqnarray}
  \label{eq:4}
  J<h<2J,
\end{eqnarray}
Here $2J>h$ ensures that there are $\hat z$-spins equal to $-1$
[otherwise we are back to (\ref{eq:9})], while $h>J$ means that two
spins joined by a positive bound are both equal to $1$. To verify it
consider
\begin{eqnarray}
  \label{eq:a18}
&& \uparrow ~ - ~ \uparrow ~ + ~ \uparrow ~ - ~ \uparrow , \\ 
&& \uparrow ~ - ~ \downarrow ~ + ~ \downarrow ~ - ~ \uparrow , 
  \label{eq:a118}
\end{eqnarray}
where $z_i=1$ and $z_i=-1$ are represented as $\uparrow$ and
$\downarrow$, respectively, while $\pm$ refer to $\tau_i =\pm$.

The energy of (\ref{eq:a18}) is $J-4h$ which is smaller than the
energy $-3J$ of (\ref{eq:a118}) precisely when $h>J$. Hence, the
minimal size of the wall in the regime (\ref{eq:4}) is just two ($+1$)
spins joined by a positive bond.

In the regime (\ref{eq:4}) there are the following domains
\cite{williams}.

\begin{itemize}
\item A frustrated $2n+1$-domains consists of an odd (and larger
than one) number $2n+1$ of consequitive minus bonds (i.e. bonds with
$\tau_k=-1$) that are bound from both ends by positive bonds.

Frustrated means the correspondence between $\{\tau_k\}$ and
$\{\hat{z}_k\}$ is not unique, e.g. the following two configurations
have the same energy
\begin{eqnarray}
  \label{eq:8}
&& + ~ \uparrow ~ - ~ \uparrow ~ - ~ \downarrow ~ - ~ \uparrow ~+, \\ 
&& + ~ \uparrow ~ - ~ \downarrow ~ - ~ \uparrow ~ - ~ \uparrow ~+. 
  \label{eq:88}
\end{eqnarray}

Both (\ref{eq:8}) and (\ref{eq:88}) have the same values of
$\{\tau_k\}_{k=1}^5$. However, the sequences $\{\hat z_k\}_{k=1}^4$ are
different, though they have exactly the same energy. More generally, a
frustrated $2n+1$-domain supports $n+1$ equivalent sequences
\cite{williams}; see also below. This non-uniqueness is reflected via
a zero-temperature entropy \cite{williams,armen_aram}.

\item Non-frustrated domains consist of an even number of
consecutive minus bonds bound from both ends by positive bonds.  In this case, 
the correspondence between $\{\tau_k\}$ and $\{\hat z_k\}$ is unique,
e.g.  $( + ~ \uparrow ~ - ~ \downarrow ~ - ~ \uparrow ~ +)$.

\end{itemize}

It is worthwhile to note that the notion of frustration described above is inherently related to the  exponential degeneracy (in the number of observations $N$) of the MAP solution above the first critical  noise intensity~\cite{armen_aram}. Indeed, consider, for instance, the frustrated 3-domain shown in~\ref{eq:a18} and~\ref{eq:a118}. In any observation sequence $\by$ of length $N$, the expected number of such frustrated 3-domains $n_d$ scales linearly with $N$, $n_d=\alpha N$, where $\alpha$ is a (possibly small) constant. And since each such domain multiplies the number of MAP solution by two (i.e., simultaneously flipping both frustrated spins is still a solution), there overall number of solutions will scale as $\propto 2^{\alpha N}$, thus indicating exponential degeneracy. 




\subsection{Further regimes}
\label{further}

Now lets us focus on the third regime defined by
\begin{eqnarray}
  \label{eq:20}
  2J/3<h<J,
\end{eqnarray}
In this regime,  the walls are formed by two  (or more) up-spins joined together by
two (or more) positive bonds. Thus $2J/3<h$ is precisely the condition
why three spins joined by two positive bonds always look up, while
$h<J$ means that two spins joined by one positive bonds can both look
down.
\begin{itemize} 
\item The old domains stay intact if they are separated (from both
ends) by (at least) three up-looking spins joined by (at least) two
positive bonds.

\item  New domains are formed by pair(s) of spins joined by a
positive bond, which are immersed into a set of negative bonds.  In
the new domains only the pairs of spins joined by positive bond can
be frustrated. The frustration rule follows the above pattern,
where now the pairs (super-spin) play the role of single spins in the
previous regime (\ref{eq:4}), while the odd (even) number of negative
spins built up into one negative (positive) superbond
\cite{williams}. Here is an example of this situation
\begin{eqnarray}
  \label{eq:21}
\underbrace{+\ua+\ua}_{\ua} 
\underbrace{ - \da - \ua -}_{-} \underbrace{\da + \da }_{\da}
\underbrace{-}_{-} \underbrace{\ua+\ua+}_{\ua},
\end{eqnarray}
where the super-spins and super-bonds are shown in the second line.
This domain is not frustrated. Note that in the regime (\ref{eq:4}),
(\ref{eq:21}) breaks down into two separate domains, the first of
which is frustrated.
\end{itemize}
The domain structure for $h<2J/3$ (and smaller values of $h$) is
constructed recursively following the above pattern \cite{williams}.

\section{Active estimation of basic domains}
\label{sec:active}

We implement active estimation for separate domains, i.e. in each
domain we supervise one spin $z_1$. Advantages of this implementation
are as follows: {\it (i)} properly choosing already one spin (and
finding out that it is incorrect) allows to get rid of the
non-uniqueness of the MAP estimate; {\it (ii)} the re-estimation $\hat
\bz(\btau)\to \tilde \bz (x_1,\btau)$ is restricted to one domain
only; {\it (iii)} (\ref{eq:7}) can be calculated approximately.

\subsection{Frustrated 3-domain}
\label{3d}

In the domain (\ref{eq:8}) we supervise the second spin. If it is
opposite to the MAP estimate, then  we move to the configuration (\ref{eq:88}) that has the
same energy:
\begin{eqnarray}
  \label{eq:u8}
&& \bhz(\btau) 
=(+ ~ \uparrow ~ - ~ \Uparrow ~ - ~ \downarrow ~ - ~ \uparrow ~+), 
 \\ 
&& \btz(\btau) 
=(+ ~ \uparrow ~ - ~ \Downarrow ~ - ~ \uparrow ~ - ~ \uparrow
~+),
  \label{eq:u88}
\end{eqnarray}
where $\Uparrow$ shows the supervised spin. Hence, the active
estimation removes the non-uniqueness of the MAP solution.  The  overlap gain is calculated from (\ref{eq:7}):
\begin{eqnarray}
  \label{eq:10}
\Delta  O=2{\sum}_{z_3}z_3 p(-1,z_3|\btau),
\end{eqnarray}
where $z_3$ is the third spin in (\ref{eq:u8}, \ref{eq:u88}); the one
that changes after the re-estimation. The marginal probability
$p(-1,z_3|\btau)$ in (\ref{eq:10}) cannnot be determined in a closed
analytic form \cite{armen_aram,hadler}.

\comment{The approximation does not apply in the vicinity of the point
  (e.g. $h=J$) where thin (two-spin) walls are broken and the domains
  are reorganized. }

We calculate $p(z_2,z_3|\btau)$ in a {\it impenetrable wall
  approximation}: when $J$ and $h$ are sufficiently large one can
neglect fluctuations of the spins within the walls separating domains
as compared to spins located within the domains. Hence, we write for
$n$-domain ($n=3,5,7,...$ and normalization is omitted)
\begin{gather}
  p(z_2,...,z_n|\btau)\propto 
  e^{-J(z_2+z_n)-J\sum_{k=2}^{n-1} z_k z_{k+1} +h\sum_{k=2}^{n}z_k }.
  \label{go}
  \end{gather}
  We checked numerically that even for thinnest two-spin walls the
  approximation works well already for $J\simeq 1$; see below.  Using
  (\ref{go}) with $n=3$ in (\ref{eq:10}) we get
\begin{gather}
  \label{eq:13}
\Delta  O=
\frac{2(1-e^{-2h})}{2+e^{-2h}+e^{-4J+2h}}.
\end{gather}
Likewise, if the originally estimate sequence is given by
(\ref{eq:v8}), then upon supervising the second spin we obtain
\begin{eqnarray}
  \label{eq:v8} 
&& \hat z(\btau) 
=(+ ~ \uparrow ~ - ~ \Downarrow ~ - ~ \uparrow ~ - ~ \uparrow ~+),
 \\ 
&& \tilde z(\btau) 
=(+ ~ \uparrow ~ - ~ \Uparrow ~ - ~ \downarrow ~ - ~ \uparrow ~+).
  \label{eq:v88}
\end{eqnarray}
Instead of (\ref{eq:10}), we need to use
\begin{eqnarray}
  \label{eq:101}
\Delta  O=-2{\sum}_{z_3}z_3 p(1,z_3|\btau),
\end{eqnarray}
which yields
\begin{eqnarray}
  \label{eq:v13}
\Delta  O=\frac{2(1-e^{-2(2J-h)})}{2+e^{-2h}+e^{-4J+2h}}.
\end{eqnarray}
Remarkably, (\ref{eq:v13}) and (\ref{eq:13}) are different. This seems
surprising, since given the apparent symmetry between (\ref{eq:u8},
\ref{eq:u88}) and (\ref{eq:v8}, \ref{eq:v88}), one would expect that
the overlap gain will be the same in both cases. This is not the case,
because (\ref{eq:101}) with (\ref{eq:10}) are defined quite
differently, e.g.  (\ref{eq:101}) involves $p(1,z_3|\btau)$, while
(\ref{eq:10}) has $p(-1,z_3|\btau)$.

Both (\ref{eq:13}) and (\ref{eq:v13}) are positive in the regime
(\ref{eq:4}), but (\ref{eq:13})$>$(\ref{eq:v13}): when the supervised
(initially) agrees with the observations, the gain of active
estimation is larger.

For $h<J$ we can employ the same two domains (\ref{eq:u8},
\ref{eq:u88}) and (\ref{eq:v8}, \ref{eq:v88}) provided that they are
surrounded by sufficiently thick walls; see Section \ref{further}. Now
the opposite relation holds: (\ref{eq:13})$<$(\ref{eq:v13}).

Note finally that whenever the supervising shows that the true value
of the spin was already recognized correctly, the overlap does not
change, but it is still useful, since the number of solutions
decreases, i.e. instead of two solutions in (\ref{eq:8}, \ref{eq:88})
we have only one. 

\comment{According to (\ref{go}) it is likely to have it
recognized correctly, because $p(z_2=1)\propto 1+e^{-4J}$,
$p(z_2=-1)\propto 2$.}

\subsection{Frustrated 5-domain}
\label{5d}
There are 3 MAP sequences for the 5-domain
\begin{eqnarray}
  \label{eq:15}
\hat \bz=(+ ~ \ua ~ - ~ \ua ~ - ~ \da ~ - ~ \UA ~ - ~ \da ~ - ~\ua ~ +), \\
  \label{eq:16}
\tilde \bz=(+ ~ \ua ~ - ~ \da ~ - ~ \ua ~ - ~ \DA ~ - ~ \ua ~ - ~\ua ~ +), \\
  \label{eq:17}
(+ ~ \ua ~ - ~ \da ~ - ~ \ua ~ - ~ \ua ~ - ~ \da ~ - ~\ua ~ +).
\end{eqnarray}
The following scenarios are the most efficient ones, i.e. they lead to
the largest $\Delta O$ (under (\ref{eq:4})): {\it (i)} going from
(\ref{eq:15}) to (\ref{eq:16}) by supervising the fourth spin (denoted
by $\UA$ in (\ref{eq:15})). {\it (ii)} going from (\ref{eq:16}) to
(\ref{eq:15}) by supervising the third spin.  Both these cases have
the same overlap gain (\ref{eq:19}) due to the mirror symmetry with
respect to the center of the domain that is shared also by (\ref{go}).

The optimal overlap gain is calculated from (\ref{go})
\begin{gather}
\Delta O= 2
\sum_{z_2,z_3,z_5}(z_3+z_5-z_2)p(z_2,z_3,-1,z_5|\btau)= \nonumber\\
\frac{2 e^{4 J} \left(e^{2 h} \left(\left(3 e^{2 h}+1\right) e^{4 J}-2 e^{2 h}+e^{4
   h}-2\right)-1\right)}{\left(3 e^{2 h}+2\right) e^{2 h+8 J}+\left(2 e^{2 h}+3 e^{4
   h}+4 e^{6 h}+1\right) e^{4 J}+e^{8 h}}. \nonumber\\
  \label{eq:19}
\end{gather}
It can be shown that (\ref{eq:19}) is larger than (\ref{eq:13}), which is the best
overlap gain for a 3-domain.  In (\ref{eq:17}) there are no spins
whose supervision can lead to the maximal overlap gain.

For the 5-domain we meet a situation that was absent in the previous
3-domain case. While (\ref{eq:19}) calculates the average overlap gain
for supervising the fourth spin in (\ref{eq:15}), it is possible that
this spin was recovered by the MAP correctly, i.e. it does not flip
after supervising. Then we are left with no overlap gain and also with
uncertainty between (\ref{eq:15}) and (\ref{eq:17}), since they both
have the correct fourth spin. We can now supervise an additional spin,
so as to recover the unique original sequence. It is seen from
(\ref{eq:15}, \ref{eq:17}) that when starting from (\ref{eq:15}) there
are 2 possibilities: to supervise the second spin or the third
one. Additional 2 possibilities exist when supervising from
(\ref{eq:17}). Out of these 4 possibilities the largest overlap gain
is achieved for supervising the second spin in (\ref{eq:15}):
\begin{gather}
  \label{eq:12}
  \Delta O= 2
  \sum_{z_3,z_5}z_3 p(-1,z_3,1,z_5|\btau)=\\
  \frac{8 e^{4 h+6 J} \sinh (h) \cosh (h-2 J)}{\left(3 e^{2
        h}+2\right) e^{2 h+8 J}+\left(2 e^{2 h}+3 e^{4 h}+4 e^{6
        h}+1\right) e^{4 J}+e^{8 h}},\nonumber
\end{gather}
where $p(-1,z_3,1,z_5|\btau)$ shows that the fourth spin
is fixed to its correct MAP value $1$. The same $\Delta O$ as in
(\ref{eq:12}) is reached when supervising the third spin in
(\ref{eq:17}). The remaining two possibilities are inferior.

Our analysis shows that Eq.~(\ref{eq:12}) is smaller than both (\ref{eq:19})
and (\ref{eq:13}). Thus, one can turn to the situation described by
(\ref{eq:12}), only if some budget is left after supervising
3-domains; cf. the discussion after (\ref{eq:14}).

\subsection{Frustrated 7-domains}

In this case there are four MAP sequences defined as follows:
\begin{eqnarray}
  \label{eq:s114}
&&\hat  \bz =(+ ~ \ua- \ua- \da- \UA- \da- \ua- \da- \ua ~ +), ~~~~~\\
  \label{eq:s115}
&&\tilde\bz=(+ ~ \ua- \da- \ua- \DA- \ua- \ua- \da- \ua ~ +),~~~~~ \\
 \label{eq:s116}
&&\hat  \bz =(+ ~ \ua -\da- \ua- \da- \UA- \da- \ua- \ua ~ +), ~~~~~\\
  \label{eq:s117}
&&\tilde\bz=(+ ~ \ua- \da- \ua- \ua- \DA- \ua- \da- \ua ~ +).~~~~~ 
\end{eqnarray}
Two most efficient active inference schemes are realized by
transitions from (\ref{eq:s114}) to (\ref{eq:s115}) and from
(\ref{eq:s116}) to (\ref{eq:s117}).  They are again related to each
other by the mirror symmetry, hence their overlap gains are equal and
is calculated from (\ref{go}). Furthermore, tedious but
straightforward calculations demonstrate that the overlap gain for
those scheme is larger than the one given by (\ref{eq:19}), the best
overlap gain for a 5-domain. We also note that  the most efficient supervision is absent if one
starts from (\ref{eq:s115}) or from (\ref{eq:s117}).

Similar to 5-domain, it is possible that a spin does not flip after semi-supervising (e.g., it was recovered correctly in the original MAP inference). In this case, one should consider supervising other spins in the domain. The analysis of possible strategies can be performed as explained in Section~\ref{5d}.

\subsection{Non-frustrated domains}

A simple analysis suggests that it is meaningless to supervise a spin inside a
non-frustrated domain (even number of negative bonds surrounded by two
walls). Indeed, consider
\begin{eqnarray}
  \label{eq:22}
&&\hat \bz   = (+ ~ \ua- \da- \UA- \da- \ua ~ +), \\
&&\tilde \bz = (+ ~ \ua- \ua- \DA- \ua- \ua ~ +), 
\end{eqnarray}
where the third spin is supervised. Calculating the overlap gain
following to the above method, we confirm that thesis gain is zero.

\subsection{Supervising inside a wall}

It is possible  that even after supervising all the frustrated domains, one still has {\em budget} for supervising additional spins. Hence, we study supervising of a spin inside a
wall. Consider the thinnest wall (two spins joined by a positive bond)
in the regime (\ref{eq:4}), and assume that it is surrounded by larger
walls (at least two positive bonds).
\begin{eqnarray}
  \label{eq:k8}
  && \bhz
  =(+\ua + ~ \uparrow ~ - ~ \Uparrow ~ + ~ \uparrow ~ - ~ \uparrow ~ +
  \ua+), ~~~~
  \\ 
  && \btz
  =(+\ua + ~ \uparrow ~ - ~ \Downarrow ~ + ~ \downarrow ~ - ~ \uparrow
  ~+\ua +),~~~~ 
  \label{eq:k88}
\end{eqnarray}
where, as above, $\Uparrow, \Downarrow$ indicate on the supervised
spin. We have [cf. (\ref{eq:10})]
\begin{eqnarray}
  \label{eq:k10}
\Delta  O=-2{\sum}_{z_4}z_4 p(-1,z_4|\btau).
\end{eqnarray}
We employ [cf. (\ref{go})] 
\begin{eqnarray}
  \label{eq:11}
p(z_3,z_4|\btau)\propto e^{-Jz_3+Jz_3z_4-Jz_4+h(z_3+z_4)}
\end{eqnarray}
and get for (\ref{eq:k10})
\begin{eqnarray}
  \label{eq:14}
\Delta    O
=\frac{2(e^{2(2J-h)}-1)}{2+e^{2h}+e^{4J-2h}}.
\end{eqnarray}
This is smaller than both (\ref{eq:13}) and (\ref{eq:v13}): it is less
efficient to supervise a spin inside a (two-spin) wall than inside a
(two-spin) domain. In the regime (\ref{eq:4}), (\ref{eq:14}) can be
both smaller or larger than (\ref{eq:12}) depending on concrete values
of $h$ and $J$.

\comment{
In the regime (\ref{eq:20}) the MAP configuration corresponds to
(\ref{eq:k88}). Hence, the active estimation in that regime works
opposite to (\ref{eq:k8}, \ref{eq:k88}): $\bhz=(\ref{eq:k88})$,
$\btz=(\ref{eq:k8})$. We then get for this regime
[cf. (\ref{eq:14})]
\begin{eqnarray}
  \label{eq:114}
\Delta    O
=\frac{2(e^{2h}-1)}{2+e^{2h}+e^{4J-2h}}.
\end{eqnarray}
This is again smaller than both (\ref{eq:13}) and (\ref{eq:v13})
(provided that (\ref{eq:13}) and (\ref{eq:v13}) in the regime
(\ref{eq:20}) refer to domains separated by 3-spin walls at
least). 
}

\subsection{Summary}

To summarize this section, we converge to the following picture of active estimation: given the
unsupervised MAP-sequence, one fixes the walls and factorizes the
sequence over frustrated domains of odd-numbered bonds. Starting from the longest domain (as it corresponds to the largest overlap gain) 
 one supervises the spin (or spins) in each domain that yield maximum expected overlap gain, as described above. Thereby, a unique supervised MAP-sequence
is gained. Once all the domains are supervised in this way, one should spend the remaining ``budget" for supervising spins inside the walls.

\section{Comparison with numerical results}
\label{sec:numerics}
We now report on our experiments aimed at validating our analytical predictions. We will focus on the non-trivial regime of sufficiently high noise intensities ($\epsilon > \epsilon_c^1)$, for which there are exponentially many (in the number of observations) MAP solutions corresponding to a given observation sequence~\cite{armen_aram}. Recalling our discussion of the frustrated domains, it is clear that all those solutions correspond to permuted  configurations of the frustrated spins. 

In our experiments, we generate random HMM sequences, then run the
Viterbi algorithm to find one of the MAP sequences. We then apply the active inference strategies described
above, and compare the numerical findings with the analytic
predictions. For simplicity, here we  focus on our prediction for
3-domains; see Section \ref{3d}. Recall that our analytical results  are
obtained under the impenetrable wall approximation (\ref{go}), which
requires that the constant $J$ is not very small;
cf. (\ref{kaskad}). In the experiments below we set $J=1.5$. 

\begin{figure*}[!t]
\centering
\subfigure[]{
    \includegraphics[width=0.65\columnwidth]{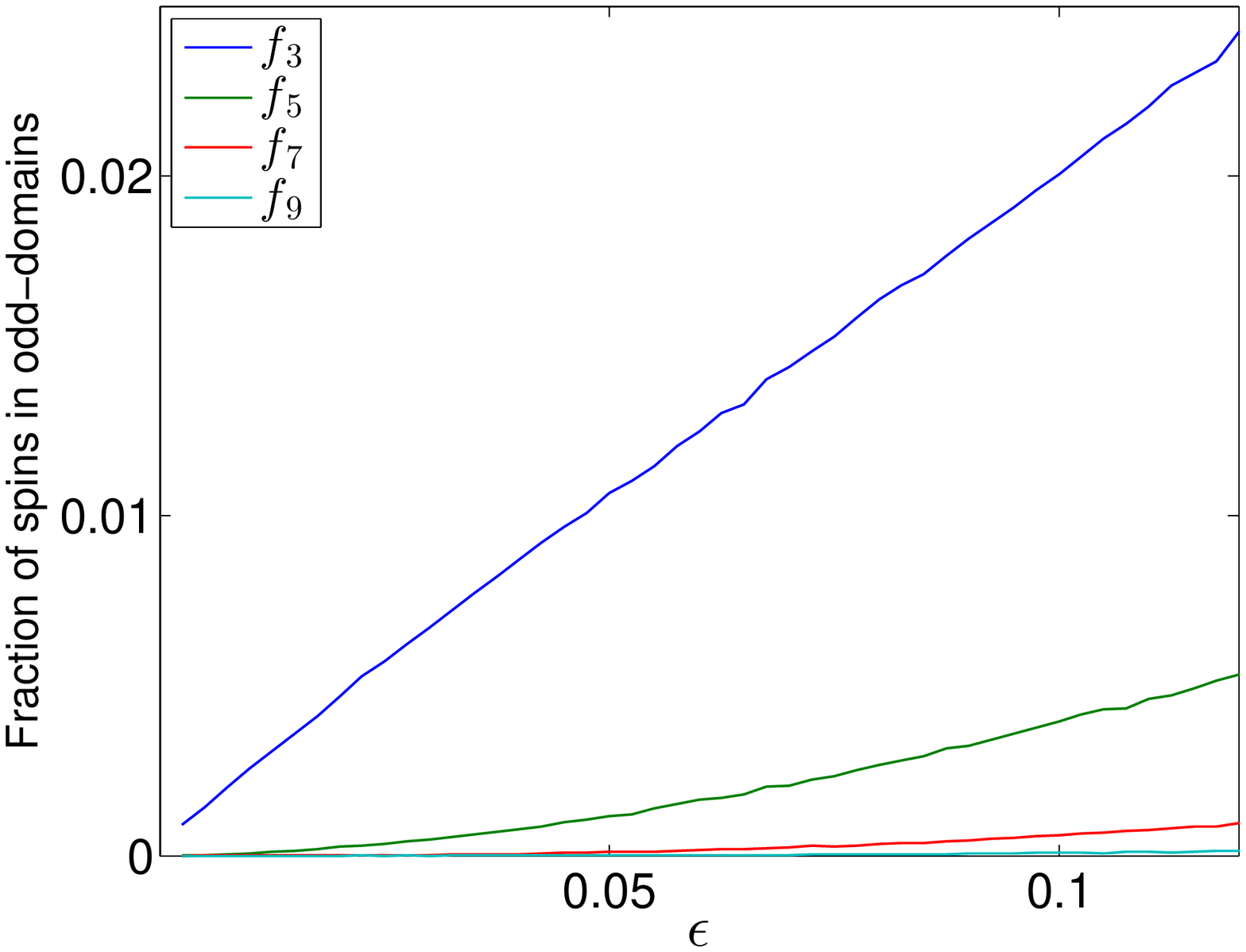} 
\label{fig:statistics}
    } 
    \subfigure[]{
    \includegraphics[width=0.65\columnwidth]{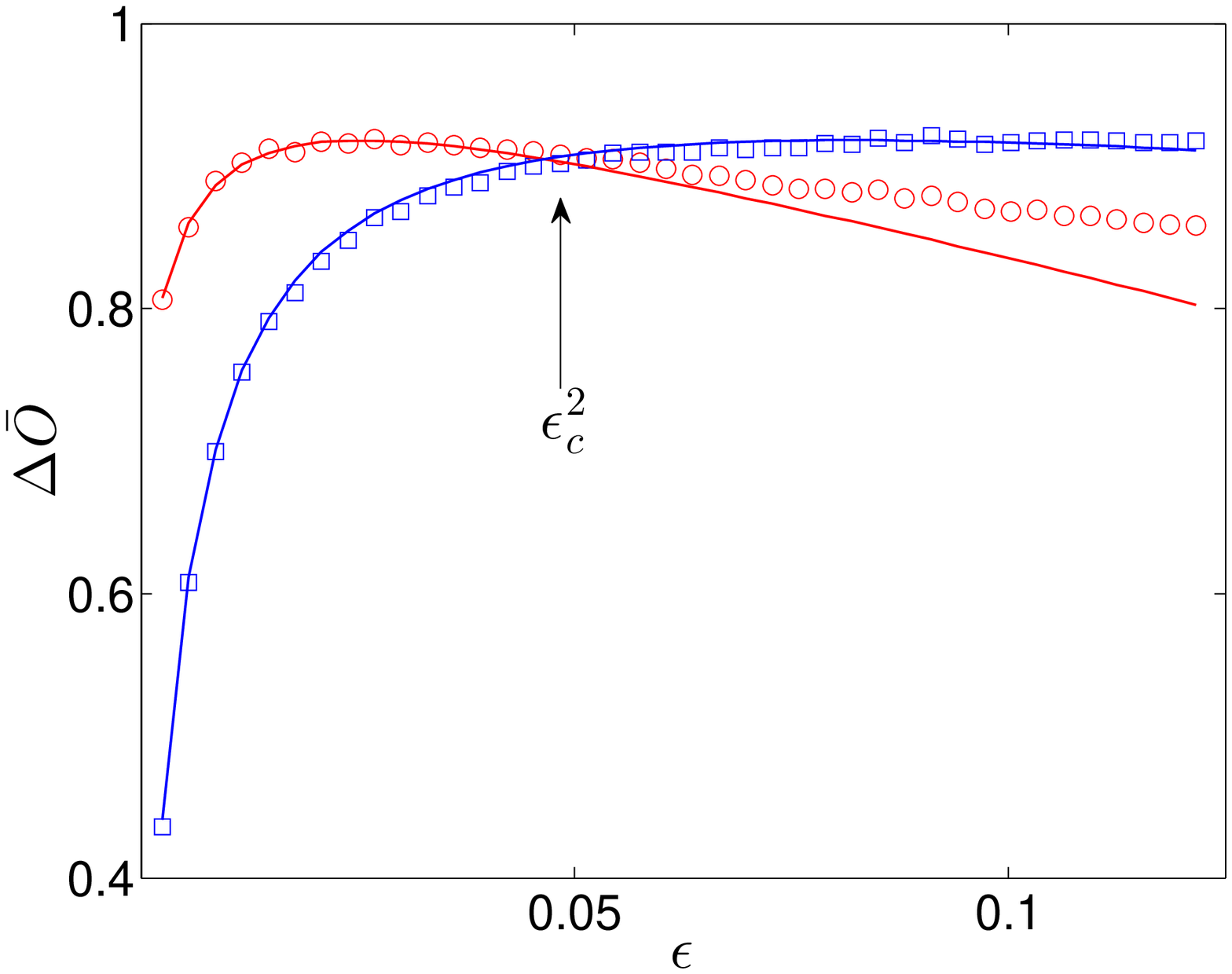}\label{fig:overlap} 
    }
    \subfigure[]{
    \includegraphics[width=0.65\columnwidth]{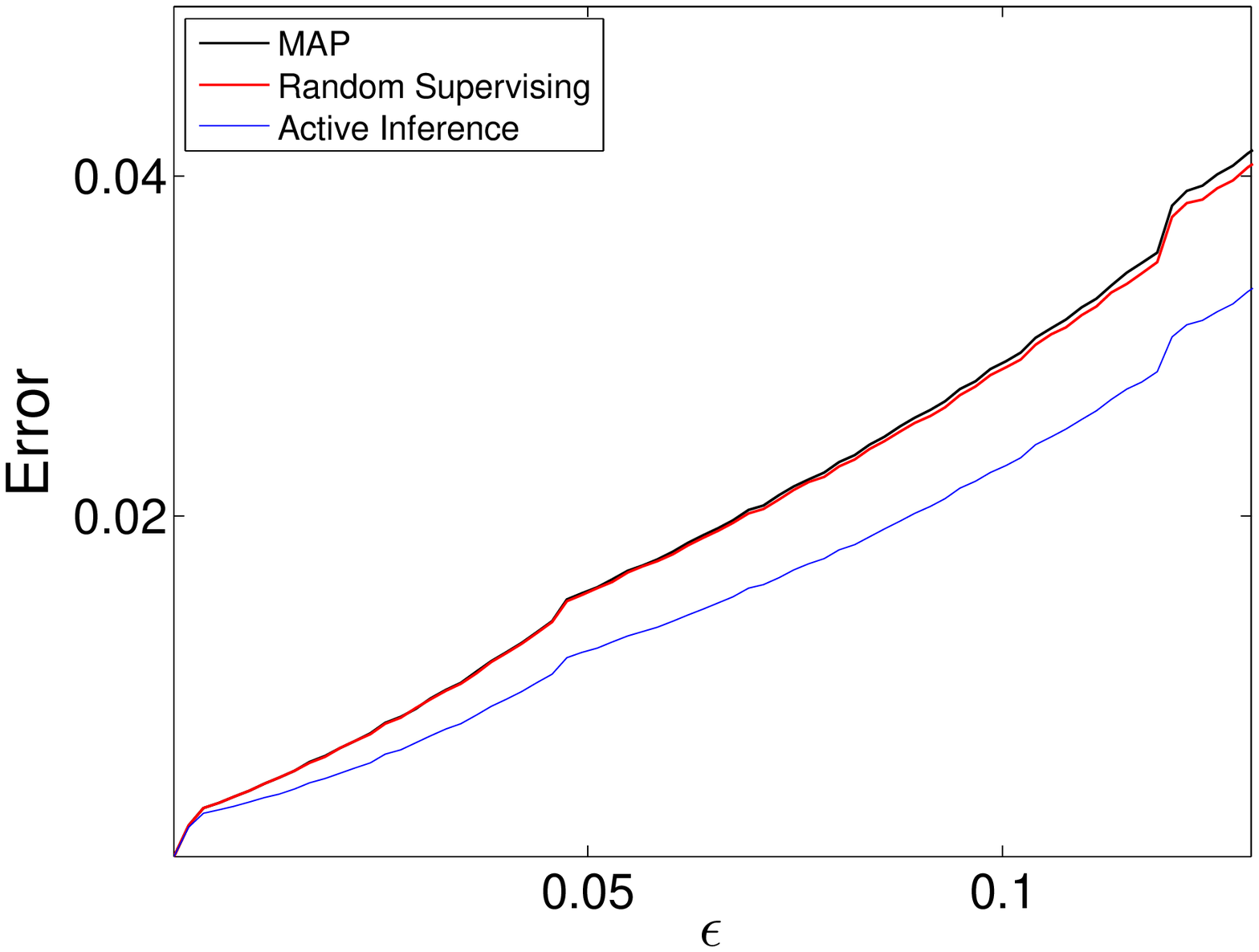}\label{fig:error} 
    }
    \caption{(a) Fraction of spins belonging to different domains plotted against noise;  (b) The average overlap gain  plotted against noise obtained from simulations (open symbols) and analytical predictions  given by Eqs.~\ref{eq:13} and ~\ref{eq:v13} (solid lines);  (c) Inference error for different methods plotted
      against  noise. In all three figures we use $J=1.5$. For (b) and (c),  we used sequences
      of size $10^5$, and averaged the results over $1000$ random realizations.  }
\end{figure*}

First, we study the statistics of different domains as a function of noise. Figure~\ref{fig:statistics} shows the fraction $f_i$ of spins that belong to frustrated domains of size  $i$, for $i=3,5,7,9$. We observe that the overall fraction of spins inside those domains is rather small. For instance, for $\epsilon=0.1$, only $2\%$ of spins belong to 3-domains, and less then $0.01\%$ belong to  9-domains. Thus, for the parameter range considered here, any gain in active inference will be mostly due to 3, 5, and 7-domains. 

%

Next, we compare our analytical prediction with simulation results  for two active inference strategies applied to 3-domains. For the simulations, we generated sufficiently long sequences ($10^5$), identified all 3-domains, and applied two active inference strategies as described in Section~\ref{3d}. For the first strategy, we always supervise one of two spins that is looking up (i.e., the inferred hidden state is aligned with the observation). The expect gain in overlap for this strategy is given by Eq.~\ref{eq:13}. And for the second strategy, we supervise the spin that is looking down (i.e., the inferred hidden state is misaligned with the observation). In this case, the expected overlap gain is given by Eq.~\ref{eq:v13}. 


The results are shown in Figure~\ref{fig:overlap}, where we  plot the expected overlap gain as a
function of noise, in the range $\epsilon_c{^1}\le \epsilon \le \epsilon_c^{3}$; 
  cf. (\ref{eq:6.b}). The agreement between the prediction and the
simulations is near-perfect when the noise intensity is in the range $\epsilon_c{^1}\le \epsilon \le \epsilon_c^{2}$. In particular, the switching of the
optimal supervising strategy, as predicted from Eqs.~(\ref{eq:13}) and
~(\ref{eq:v13}), is reproduced in the experiments. Interestingly, for one of the ``branches" (given by Eq.~\ref{eq:v13}), the agreement remains near-perfect even for larger noise intensities. However, the agreement with the other branch (given by Eq.~\ref{eq:v13}) deteriorates with increasing noise. We recall that the analytical predictions are obtained under the impenetrable wall approximation, which assumes that domains are surrounded by sufficiently {\em thick} walls. When increasing noise, the assumptions starts to gradually break down, and the approximation becomes less accurate. 


\comment{Recall that the analysis in section \ref{3d} is carried out in the
parameter range $J<h<2J$, or correspondingly, . The numerical data for different values of
$J$ is represented in a uniform way, if instead of $h$ we employ (in
the regime (\ref{eq:4})) the normalized noise
$\frac{\epsilon-\epsilon_c^{1}}{\epsilon_c^{2}-\epsilon_c^{1}}$, where
$\epsilon_c^{(m)}\equiv 1/[1+\e^{4J/m}]$ for $m=1,2,..$
\cite{armen_aram}. Then (\ref{eq:4}) is written as
$\epsilon_c^{(1)}<\epsilon<\epsilon_c^{(2)}$. }

Finally, Figure~\ref{fig:error} compares inference errors for the original MAP, MAP with random
supervising, and MAP with active inference. We see that the
active inference strategy yields lower inference error over the whole range
of noise intensities. In fact, this difference becomes more pronounced
for large noise intensities. We also note that the slight {\em kinks} in the curves corresponding to the phase transition between different domains~\cite{armen_aram}.



\section{Conclusion}
\label{sec:conclusion}

\subsection{Discussion of the main results} We have developed an
analytical approach to study active inference problem in binary
symmetric HMMs. We obtained a closed-form expression that relates the
expected accuracy gain to model parameters within the impenetrable
domain approximation, specified the validity range of the analysis,
and tested our analytical predictions with numerical
simulations. Based on the analytical insights, we suggest an optimal
active inference strategies for BS-HMM, that can be summarized as
follows: Given an observation sequence $\by$ and a corresponding
(unsupervised) MAP-sequence $\hat{\bx}(\by)$, we first find all the
frustrated domains composed of odd-numbered bonds. Then, starting from
the longer domains, we supervise one of the spins inside the domain,
according to the optimality criteria outline in (\ref{eq:13},
\ref{eq:19}). Only after supervising all the frustrated domains (and
subject to budget constraints), one can consider supervising schemes
described in (\ref{eq:12}) and (\ref{eq:14}).

Note that while our focus was on batch active inference, our results
are also applicable to the online settings (e.g., active filtering),
owing to the separation of the estimated hidden sequence into separate
weakly interacting domains.

\subsection{Relationship to existing heuristics for active inference}
Here we discuss to which extent solutions studied above relate to
heuristic methods for active inference; see
Section~\ref{sec:intro}. Below we always assume the regime of
parameters, where (\ref{go})
applies. 

Within the {\it uncertainty reduction} heuristics we supervise $x_l$
if some suitable measure of uncertainty, e.g. $1-\sum_{x_k=\pm 1}
p^2(x_k|\by)$, maximizes at $k=l$. For the 3-domain (\ref{eq:u8},
\ref{eq:u88}) both spins of the domain are equally uncertain
[see (\ref{go})], but we saw that their supervising leads to
different results. Thus, the uncertainty reduction does not
correlate with the optimal scheme.

For the 5-domain (\ref{eq:15}--\ref{eq:17}) the third spin (and due to
the mirror symmetry the fourth one) is the most uncertain one. This
correlates with the optimal scheme (\ref{eq:15}-\ref{eq:16}),
but does not allow to recover this scheme, e.g. supervising the third
spin in (\ref{eq:17}) is not optimal. The same correlation is
seen for the 7-domain, where the fourth spin is the most uncertain
one; see (\ref{eq:s114}--\ref{eq:s117}).

Consider now another (related) heuristic principle for active
inference that tries to supervise spins in such a way that it leads to
a unique solution. This principle does not apply 3-domain, since here
the choice of either spin leads to a unique solution. It partially
holds for the optimal solution of the 5-domain: if in (\ref{eq:15})
the supervised spin was found to be wrong, then we are left with only
one possible configuration (\ref{eq:16}), e.g. supervising the third
spin in (\ref{eq:15}) does lead to unique sequence: it can be either
(\ref{eq:16}) or (\ref{eq:17}). However, if the supervised spin in
(\ref{eq:15}) was found to be correct, we are not left with a unique
configuration [an additional supervising is necessary to achieve a
unique configuration, but it is beneficial to supervise instead
another 3-domain of 5-domain; see the discussion after
(\ref{eq:12})]. Likewise, for the 7-domain, no correlation exists
between the optimal supervising and the uniqueness of the resulting
configuration: even if the optimally supervised is found to be wrong,
we do not automatically appear in a unique configuration; see
(\ref{eq:s114}, \ref{eq:s115}).

\subsection{Open Problems} There are several interesting directions
for extending the presented results. First, here we focused on the
inference task, assuming that the model parameters are known. In a
more realistic scenario, those parameters ($(q,\epsilon)$, or $(J,h)$)
are unknown and need to be estimated from the data, e.g., using
Baum-Welch algorithm~\cite{rabiner_review}. While developing a full
analytical solution for the joint active inference/learning problem
might be challenging even for the simple BS-HMM considered here,
it should be possible to extend the methods presented here for
studying the robustness of optimal inference strategies with respect
to misspecified parameters.

Another interesting and important question is to what extent the
optimal active inference schemes analyzed here apply to more general
class of HMMs. We believe that the answer to this question can be
examined via a generalized definition of frustration. In the context
of MAP estimation, this implies looking at generalized (k-best)
Viterbi algorithm that return several candidates solutions having
approximately equal likelihoods or energies~\cite{foreman,ep}. In this
scenario , we can intuitively define frustrated variables as those
that change from one solution to another.

Also, using the analogy with statistical physics, we believe that the
active MAP inference via domains and walls can be generalized beyond
the (one-dimensional) HMM models, and it possibly applies to active
recognition of two-dimensional patterns.  Now instead of
one-dimensional Ising model, we get a two-dimensional random Ising
spin system. Its $T=0$ situation corresponds to the MAP recognition of
patterns.  We note that two-dimensional Ising-type models haven been
used extensively in various pattern recognition tasks such as computer
vision~\cite{geman}.

Finally, there are several interesting analogies between the
optimal inference strategies uncovered here and human heuristics in
information processing \cite{psy,plos}. Note that the optimal scheme
in the regime (\ref{eq:4}) amounts to
supervising the up-spins, e.g., the ones that agree with observations.
This point can be related to {\it falsifiability}
\cite{psy}, where one looks at cases when the existing (prior) knowledge
can be challenged. In contrast, people often show {\it
  confirmation bias}, i.e. they tend to confirm the existing knowledge
rather than  question it; see \cite{plos} for a review.  Further
development of those ideas may provide interesting connections with cognitive aspects of active
inference strategies.

\end{document}